%% file: main_tex_arxiv.tex
\documentclass[10pt,twocolumn,letterpaper]{article}

\usepackage[]{cvpr}      
\usepackage{multirow}
\usepackage[T1]{fontenc}

\input{preamble}

\definecolor{cvprblue}{rgb}{0.21,0.49,0.74}
\usepackage[pagebackref,breaklinks,colorlinks,allcolors=cvprblue]{hyperref}

\def\paperID{8434} 
\def\confName{CVPR}
\def\confYear{2026}

\title{DyStream: Streaming Dyadic Talking Heads Generation via Flow Matching-based Autoregressive Model}

\author{Bohong Chen\textsuperscript{1}\quad 
Haiyang Liu\textsuperscript{2} \quad 
\\
\textsuperscript{1}Zhejiang University \quad
\textsuperscript{2}The University of Tokyo \quad \\
\\
\url{https://robinwitch.github.io/DyStream-Page}
}

\newcommand{\ourmethod}{DyStream\xspace}

\begin{document}
\maketitle

\begin{abstract}
Generating realistic, dyadic talking head video requires ultra-low latency. Existing chunk-based methods require full non-causal context windows, introducing significant delays. This high latency critically prevents the immediate, non-verbal feedback required for a realistic listener.
To address this, we present \ourmethod, a flow matching-based autoregressive model that could generate video in real-time from both speaker and listener audio.
Our method contains two key designs: (1) we adopt a stream-friendly autoregressive framework with flow-matching heads for probabilistic modeling, and (2) We propose a causal encoder enhanced by a lookahead module to incorporate short future context (\textit{e.g.}, 60\,ms)  to improve quality while maintaining low latency.  
Our analysis shows this simple-and-effective method significantly surpass alternative causal strategies, including distillation and generative encoder. Extensive experiments show that \ourmethod could generate video within 34\,ms per frame, guaranteeing the entire system latency remains under 100\,ms. Besides, it achieves state-of-the-art lip-sync quality, with offline and online LipSync Confidence scores of 8.13 and 7.61 on HDTF, respectively. 
The model, weights and codes are available.  
\end{abstract}

\section{Introduction}
\label{sec.intro}
Recent breakthroughs in full-duplex speech generation systems \citep{wang2024full,defossez2024moshi} have paved the way for highly interactive, streaming AI agents. However, these speech-only agents lack a visual presence, limiting their application in areas like education, sales, and virtual companionship. Bridging this gap requires a model that can generate real-time talking-head video from a single image and the streaming dual-track audio source.
\begin{figure}
    \centering
    \includegraphics[width=\linewidth]{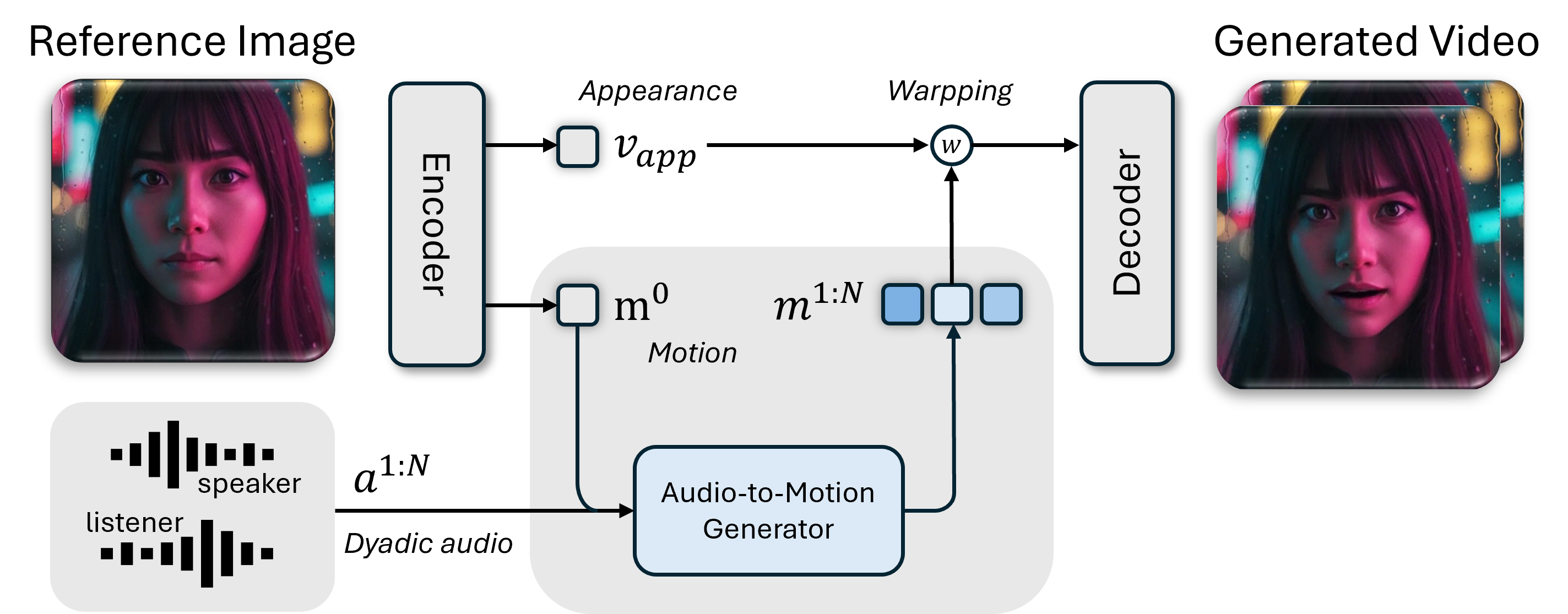}
    \caption{\textbf{System pipeline.} 
    \ourmethod generates talking-head videos from a single reference image and dyadic stream. 
    First, an Autoencoder disentangles the reference image into a static appearance feature $\mathbf{v}_{app}$ and an initial, identity-agnostic motion feature $\mathbf{m}^0$. 
    Next, the Audio-to-Motion Generator takes the initial motion $\mathbf{m}^0$ and the audio stream as input to generate a new sequence of audio-aligned motion features $\mathbf{m}^{1:N}$. 
    Finally, the Autoencoder's decoder synthesizes the output video by warping the appearance feature $\mathbf{v}_{app}$ according to the generated motion sequence $\mathbf{m}^{1:N}$.
    }   
    \label{fig:overview1}
\end{figure}

Current state-of-the-art approaches~\citep{zhu2025infp} have primarily focused on exploring offline solutions. These systems, typically based on two-stage, chunk-by-chunk models, achieve good generation quality. However, their design requires the entire audio chunk's content before generation begin, introduces a fundamental latency problem. This accumulated latency, often hundreds of milliseconds, is acceptable when the agent is speaking but becomes a critical bottleneck when it needs to act as an active listener. A listener must react instantaneously to the user's speech with subtle non-verbal cues like nods and facial expressions to create a sense of engagement. The inherent delay in chunk-based systems limits their performance on immediate, naturalistic responses.

To address the latency issue, we propose a new flow-matching based autoregressive model, \ourmethod. It accepts a streaming, dual-track audio input—capturing both the user's speech and the agent's own generated speech—and generates the corresponding talking head video frame-by-frame. This autoregressive (AR) architecture fundamentally reduces the latency bottleneck and could generate a listener response with one-token latency in theory.

However, achieving high-fidelity lip synchronization in this causal, frame-by-frame manner is challenging. As human speech involves anticipatory coarticulation, where mouth movements prepare for upcoming phonemes slightly before the sound is produced. This makes it difficult for a purely causal model, without future information, to generate accurate lip shapes. To solve this, we design a causal audio encoder with a minimal audio lookahead module. Specifically, the model generates the current video frame by conditioning on an extremely short segment of future audio (\textit{e.g.}, 60ms). Crucially, this lookahead duration is designed to be shorter than the typical packet size of streaming audio APIs (\textit{e.g.}, ~100ms from the OpenAI Realtime API). This ensures that our system can process an entire incoming audio packet and generate the corresponding video before the next packet arrives, thus achieving system-level real-time performance and operating seamlessly within a streaming infrastructure. In addition, we analyzed alternative causal encoder designs, such as distillation and generative encoders, and found the lookahead module to be simple yet effective.

With the above designs, our system achieves state-of-the-art lip-sync quality, with offline and online LipSync Confidence scores of 8.13 and 7.61 on HDTF, respectively. Furthermore, it ensures an immediate and fluid interactive experience by processing any incoming audio packet within 40ms. This level of responsiveness makes our model suitable for creating interactive visual agents that can not only speak but also listen. Our contributions can be summarized as follows:

\begin{itemize}
    \item We propose a new flow-matching based autoregressive model that generates talking head videos frame-by-frame, theoretically reducing the generation latency compared to previous chunk-based methods.
    \item We introduce a causal audio encoder with a minimal audio lookahead mechanism (\textit{e.g.}, 60ms) and demonstrate it is simple yet effective for enabling high-quality lip-sync while maintaining system-level real-time responsiveness.
    \item Our method achieves state-of-the-art performance in lip-sync accuracy and guarantees a generation latency under 34ms.
\end{itemize}

\section{Related Work}
\label{sec.related_work}
\paragraph{Video Diffusion-based Talking Heads Generation}
A prominent approach involves finetuning a pre-trained image-to-video (I2V) model to align the generated video content with an audio condition, mainly for achieving accurate lip synchronization~\citep{tian2024emo,xu2024hallo,jiang2024loopy,lin2025omnihuman,chen2025hunyuanvideo_avatar,wei2025mocha,zhou2025gohd,bigata2025keysync,li2025ditto,hong2025audio,zhong2025FADA,guan2025audcast,yi2025magicinfinite,yang2025megactor,liu2024tango,liu2025video,qin2025versatile,bigata2025keyface,mu2025flap,sun2024uniavatar,wang2025emotivetalk,chen2025echomimic}.
Early works in this domain were often based on U-Net architectures from Stable Diffusion~\citep{rombach2022high}. For instance, EMO~\citep{tian2024emo} integrates audio conditions via an additional cross-attention layer, while Loopy~\citep{jiang2024loopy} designs a past-frame compression scheme to better support longer video generation. More recent works have shifted towards Transformer-based backbones (DiT), such as Hallo~\citep{xu2024hallo}, HunyuanVideo-Avatar~\citep{chen2025hunyuanvideo_avatar}, and MOCHA~\citep{wei2025mocha}, which largely follow the cross-attention mechanism to inject audio features. While all these methods demonstrate high-quality results with few artifacts and strong generalization, they suffer from high computational costs and slow inference speeds.

\paragraph{Two-Stage Warp-based Talking Heads Generation}
Two-stage methods aim to first predict disentangled appearance and motion representations from audio, and then use these representations to warp a source image~\citep{drobyshev2022megaportraits,drobyshev2024emoportraits,xu2024vasa,zhu2025infp,lu2021live,zhou2020makelttalk,zhang2023sadtalker,wang2025omnitalker,qi2025chatanyone,huang2025se4lip,chen2025audio,gong2025MGGTalk,fu2025DAMC,shen2025Audio-Plane,liu2025disenttalk,zhen2025teller,liu2025syncanimation,guo2025arig}. Representative works such as MegaPortraits~\citep{drobyshev2022megaportraits}, VASA~\citep{xu2024vasa}, and INFP~\citep{zhu2025infp} have explored various designs for a robust, identity-agnostic facial motion space. Recent variants further improve flexibility and style controllability, including OmniTalker~\citep{wang2025omnitalker}, and ChatAnyone~\citep{qi2025chatanyone}. For the audio-to-motion stage, they typically adopt a chunk by chunk diffusion strategy. However, the requirement of the entire audio chunk’s content leads to a latency problem. Different with them, we introduce a flow-matching based autoregressive model to guarantee one token latency in theory.

\paragraph{Dyadic Talking Heads Generation}
The early approaches in dyadic talking heads generation typically treated conversations as multi-turn exchanges, where the agent acted as either a speaker or listener in each turn. MultiDialog~\cite{GoRealTalk} generated speaker behaviors conditioned on the user’s previous utterance, but failed to produce synchronized listener reactions. To jointly model both roles, ViCo-X~\cite{zhou2025interactive} proposes a Role Switcher bridging separate generators for speaker and listener, but the explicit switching often caused unnatural transitions. DIM~\cite{tran2024dim} pretrained a unified model to capture dyadic context but still required manual role assignment during adaptation. AgentAvatar~\cite{wang2023agentavatar} synthesized photo-realistic avatars driven by textual descriptions, but its motions lacked precise audio alignment.
Recent person-specific studies~\cite{sun2024beyond,ng2024Audio2Photoreal,huang2024interact,yan2023dialoguenerf} achieve high realism but perform limited on unseen identities. 
In contrast to these approaches, which often rely on explicit role-switching rules, multiple expert models, or offline processing of dual audio tracks, our method is designed for online, streaming applications. We utilize a single, unified model to generate both speaker and listener behaviors concurrently from dual-track audio inputs.

\begin{figure*}
    \centering
    \includegraphics[width=0.9\linewidth]{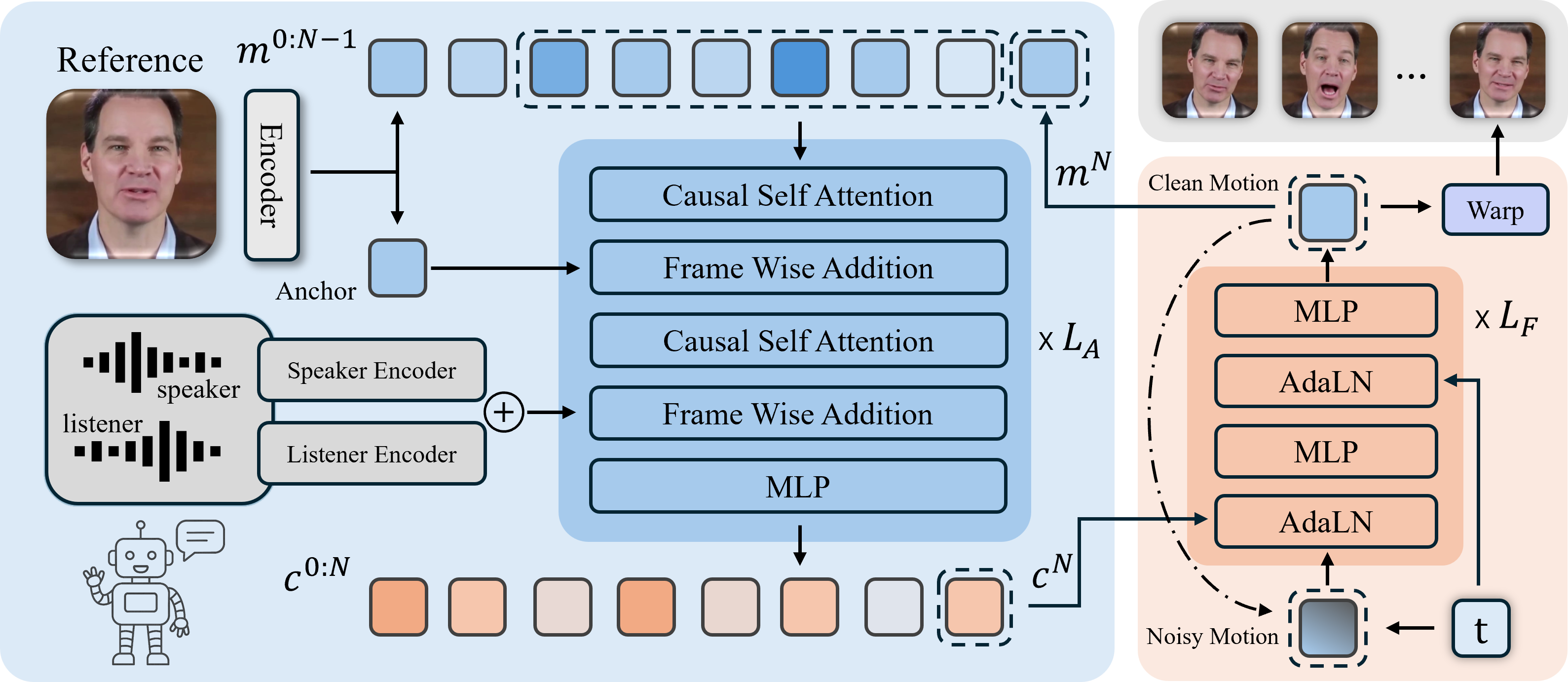}
    \caption{\textbf{The architecture of our audio-to-motion generator.} 
    Our model comprises two core modules: an autoregressive network (blue) and a flow matching head (orange). The autoregressive network, built from causal self-attention and MLP blocks, processes the audio, anchor, and previous motion inputs to generate a conditioning signal $c^N$. This signal is fed into the flow matching head, a stack of MLPs and AdaLN layers. Here, it is injected via AdaLN to guide a multi-step flow matching process to produce the final motion $m^N$. Finally, the newly generated motion $m^N$ is used to warp the reference image into the output frame, while simultaneously being fed back into the autoregressive network as input for the subsequent generation.
    }
    \label{fig:overview}
\end{figure*}

\section{Methodology}
\label{sec:method}
Given a single image and a streaming dyadic audio signal, our method generates streaming talking-head videos. It is based on a two-stage framework. The first stage disentangles appearance and motion from the source image using an off-the-shelf model based on LIA~\citep{wang2024lia}, which we briefly review in Section~\ref{sec:3.1}. The second stage implements audio-to-motion generation using a flow matching-based autoregressive model, as detailed in Section~\ref{sec:3.2}. Furthermore, to facilitate real-time responsiveness, we design a causal audio encoder with a lookahead module (Section~\ref{sec:3.3}) to minimize its reliance on future audio frames.

\subsection{Motion-aware Autoencoder}
\label{sec:3.1}
The first stage contains an Autoencoder (AE) to disentangle a source image $\mathbf{I}_s$ into two distinct representations: a static appearance feature $\mathbf{v}_{app}$ that encodes the subject's identity, and a dynamic motion feature $\mathbf{m}$ that describes the pose and expression. Conceptually, this AE includes: an appearance encoder ($\mathcal{E}_{app}$), a motion encoder ($\mathcal{E}_{m}$), a flow estimator ($\mathcal{F}$), and a final decoder ($\mathcal{D}_{vae}$). These components are trained jointly in a self-supervised manner, using a video reconstruction task that reconstruct a driving video by combining the appearance from a source image with the motion from the driving video itself. In our implementation, we adopt the architecture and training objective from LivePortrait~\citep{guo2024liveportrait} and LIA~\citep{wang2024lia}. 

First, the motion encoder $\mathcal{E}_m$ extracts a source motion feature $\mathbf{m}_s$ from the source image $\mathbf{I}_s$, and a sequence of driving motion features $\mathbf{m}^{1:N}_{dri}$ from a driving video $\mathcal{V}_{dri}$.  
\begin{equation}
\resizebox{0.9\columnwidth}{!}{$\mathbf{m}_{s} = \mathcal{E}_{m}(\mathbf{I}_s), \quad \mathbf{v}_{app} = \mathcal{E}_{app}(\mathbf{I}_s), \quad \mathbf{m}^{1:N}_{dri} = \mathcal{E}_{m}(\mathcal{V}_{dri})$}
\end{equation}
Subsequently, a motion flow estimator $\mathcal{F}$ predicts a dense flow field $\mathbf{F}_{s \rightarrow d}$ from the motion features. The decoder $\mathcal{D}_{vae}$ then synthesizes the final video frames $\hat{\mathbf{I}}^{1:N}_{pred}$ by warping the source appearance feature $\mathbf{v}_{app}$ with this flow field.
\begin{equation}
\resizebox{0.9\columnwidth}{!}{$\mathbf{F}_{s \rightarrow d} = \mathcal{F}(\mathbf{m}_{s}, \mathbf{m}^{1:N}_{dri}); \quad \hat{\mathbf{I}}^{1:N}_{pred} = \mathcal{D}_{vae}(\text{Warp}(\mathbf{v}_{app}, \mathbf{F}_{s \rightarrow d}))$}
\end{equation}
The module is trained with a composite loss function $\mathcal{L}_{AE}$ combining reconstruction, adversarial, perceptual, cycle consistency, and gaze direction losses. More details are provided in Appendix A.

\subsection{Flow matching-based Autoregressive Model}
\label{sec:3.2}
The second stage employs an Audio-Driven Motion Generation Transformer, which we denote as $\mathcal{D}_{\theta}$. This model generates a sequence of motion latents $\mathbf{m}^{1:N}$ conditioned on a corresponding dyadic audio feature sequence $\mathbf{a}^{1:N}$ and the initial motion latent $\mathbf{m}_s$ extracted from the source image.

\paragraph{Model Architecture.}
As shown in Figure~\ref{fig:overview}, the core architecture of our network is a combination of a flow matching head and an autoregressive (AR) model. Motivated by~\cite{li2024autoregressive}, the AR model regresses the continuous, deterministic feature, and a single frame-level flow matching head is for stochastic modeling.
Specifically, the feature from the AR model, along with the denoising timestep and noise, are fed as conditions for the flow matching head. The denoised clean motion latent is then fed to the AR model for next-frame generation.

\paragraph{Implementation Details.}
For the AR model, we adapt the 1D Transformer block design from Wan2.1 \citep{wan2025wan}. A key change is our use of block-independent projections for timestamp embeddings instead of the original shared projections. The flow matching head is composed of stacked MLPs.
For audio conditioning, the dyadic audio feature is from the concatenation of encoded speaker and listener audio features. We adopt a customized Wav2Vec2 \citep{baevski2020wav2vec} as an audio encoder; the details are in Section \ref{sec:3.3}. As shown in Figure \ref{fig:overview}, we leverage linear interpolation to align with the temporal dimension of the motion features, and fuse the audio feature via element-wise addition. We found that this is similar to a cross-attention with a diagonal attention mask, which shows better performance in our case.

\paragraph{Anchor Conditioning for Anti-Drifting.} 
As an inherently frame-by-frame autoregressive model, our approach inevitably faces error accumulation during long-sequence generation. The subject's head position gradually drifts from its initial location in the reference image, causing severe artifacts and a decline in visual quality. Considering that our inference procedure generates only one new frame per forward pass in the sliding window—with previously generated frames serving as historical context—we introduce a corresponding modification to the training strategy. Specifically, during training,
we randomly select one frame from the last ten frames of the sequence to serve as the appearance anchor during training. During inference, however, we consistently use only the initial reference frame as the anchor. This randomized anchor training strategy effectively mitigating pose drift during autoregressive generation.

\paragraph{Training and Inference.}
For training, we adopt the ${m}_0$-prediction objective, where the model learns to predict the clean motion latent $\mathbf{m}_0$ from its noised version $\mathbf{m}_t$. The loss function is formulated as:
\begin{equation}
    \mathcal{L}_{\text{flow}} = \mathbb{E}_{t, \boldsymbol{\epsilon}} \left[ \| \mathcal{D}_{\theta}(\mathbf{m}_t, t, \mathbf{c}) - \mathbf{m}_0 \|^2 \right]
\end{equation}
where $\mathbf{m}_t = (1-\sigma_t)\mathbf{m}_0 + \sigma_t\boldsymbol{\epsilon}$ with $t \sim \mathcal{U}[0,1]$ and $\boldsymbol{\epsilon} \sim \mathcal{N}(\mathbf{0}, \mathbf{I})$. The conditioning signal $\mathbf{c}$ is the output from AR model.

During inference, we compute the velocity field from the predicted $\hat{\mathbf{m}}_0$ and integrate using an Euler-based ODE solver. To enable fine-grained control, we employ a multi-conditional guidance strategy. Let $\hat{\mathbf{m}}(\mathbf{c}_S, \mathbf{c}_L, \mathbf{c}_R) = \mathbf{m}_{\theta}(\mathbf{m}_t, t, \mathbf{c}_S, \mathbf{c}_L, \mathbf{c}_R)$ denote the model's prediction given the speaker ($\mathbf{S}$), listener ($\mathbf{L}$), and anchor ($\mathbf{R}$) conditions. The unconditional prediction is $\hat{\mathbf{m}}_{\text{uncond}} = \hat{\mathbf{m}}(\mathbf{0}, \mathbf{0}, \mathbf{0})$.

The final guided prediction $\mathbf{m}_{\text{cond}}$ is computed by rearranging the standard CFG formulation:
\begin{equation}
\begin{split}
    \mathbf{m}_{\text{cond}} ={}& (1 - w_{\Sigma})\hat{\mathbf{m}}_{\text{uncond}} + w_s \hat{\mathbf{m}}(\mathbf{S}, \mathbf{0}, \mathbf{0}) \\
    & + w_l \hat{\mathbf{m}}(\mathbf{0}, \mathbf{L}, \mathbf{0}) + w_r \hat{\mathbf{m}}(\mathbf{0}, \mathbf{0}, \mathbf{R}) \\
    & + w_{\text{all}} \hat{\mathbf{m}}(\mathbf{S}, \mathbf{L}, \mathbf{R}),
\end{split}
\label{eq:cfg}
\end{equation}
where $w_s, w_l, w_r,$ and $w_{\text{all}}$ are the respective guidance scales, and $w_{\Sigma} = w_s + w_l + w_r + w_{\text{all}}$.

Besides, we maintain a sliding window of N frames matching the training length. Initially, the first N-1 frames are set to static motion with the corresponding dyadic audio input to predict the next frame. We then shift the window by removing the oldest frame and appending the newly generated frame, repeating this process autoregressively to generate the entire sequence frame by frame.

\subsection{Make It Realtime with Customized Wav2Vec2}
\label{sec:3.3}

While current models could generate video in a causal frame-by-frame manner in theory, their lipsync performance is limited with a straightforward implementation. A primary reason is that standard audio encoders, such as Wav2Vec2 \citep{baevski2020wav2vec}, typically require non-causal context to extract current features. This leads to severe feature degradation when causality is enforced during inference. As shown in Table~\ref{tab:wavencoder}, the model's audio-lip synchronization improves as the current audio frame can access more future audio frames. To solve this, we customize a causal Wav2Vec2 with a minimal lookahead module, restricting its receptive field to a maximum of $n$ future audio frames.

To implement this causal Wav2Vec2, we introduce two key modifications to eliminate its inherent non-causal operations. First, the standard architecture utilizes GroupNorm in its initial convolutional layer. Since GroupNorm computes statistics across the entire input sequence, it introduces a dependency on future frames. We replace it with LayerNorm, which operates independently at each timestep, thus ensuring causality.
Second, a more significant source of non-causality lies in its positional encoding, implemented as a convolution with a large kernel size of $128$. This layer inherently accesses up to $64$ future frames (a 1.28-second lookahead). We therefore remove this convolutional positional encoding entirely and instead integrate Rotary Positional Embeddings (RoPE) \citep{su2024roformer} into the Transformer attention layers.

These modifications are important to remove implicit, hard-coded lookaheads from the architecture. As a result, the model's temporal receptive field is governed solely by the attention mask. This grants us precise and explicit control over the amount of future audio context the model can access.

We then decide the number of lookahead frames based on the demands of real-time processing. For instance, systems like the OpenAI Realtime API~\cite{openai2024gpt4ocard} deliver audio in 100ms chunks. Our target video generation rate is 25 fps, which corresponds to one video frame per 40ms of audio. By configuring our modified audio encoder to access an additional 60ms or less of future audio, we can generate a video frame immediately upon receiving the first 100ms audio segment, thereby incurring no additional latency. This lookahead is implemented by adjusting the attention mask inside our customized Wav2Vec2.

In contrast, for the listener part, we adopt a different approach. The average human reaction time in a conversation is approximately 300 milliseconds~\cite{conversationlatent}. Unlike a speaker, a listener may not require future audio context to determine precise lip movements. Therefore, for the listener's audio encoder, we implement a purely causal attention mask. This ensures that the model's responses are generated based only on past and current audio information.

\paragraph{Training Details.}
Our training strategy for the Rope-Wav2Vec2 models differs based on their attention mechanism. The Rope-Wav2Vec2 model with a full attention mask is trained from scratch, using the same pre-training task as the original Wav2Vec2. Subsequently, for the variant with restricted attention, we do not train it from scratch. Instead, we employ knowledge distillation, using the pre-trained full-attention model as the teacher to train the restricted-attention model. These final distilled models are then used for our generation task.

\section{Experiments}
\subsection{Implementation Details}
\paragraph{Architecture.} Our model has three components: an Autoencoder (AE), an Autoregressive model, and a Flow matching Head, with 110M, 182M, and 80M parameters, respectively. The Autoregressive model consists of 12 blocks, each with a hidden dimension of 768 and 8 attention heads. The Flowmatching Head is composed of 6 blocks, also featuring a hidden dimension of 768.

\paragraph{Autoencoder Training.} We train the AE using the AdamW optimizer~\citep{loshchilov2017Adamw} with a learning rate of 1e-5 and a batch size of 48. The dimension of the latent motion codes is set to 512. The AE is trained for a total of 150,000 iterations on H200.

\paragraph{Autoregressive Flow Matching Training.} 
We apply a dropout of 0.5 to the speaker and listener audio, and 0.1 to the reference anchor. For inference, we employ a 5-step Euler ODE sampler and utilize Classifier-Free Guidance (CFG)~\citep{ho2022classifier}, as described in Equation~\ref{eq:cfg}, with guidance weights set to $w_{all}=1.0$, $w_{s}=0.5$, $w_{l}=0.5$, and $w_{r}=0.5$. The model is trained for 120,000 iterations using the AdamW optimizer with a learning rate of 2e-5 and a global batch size of 64, and costs 12 hours on a single H200.

\subsection{Dataset}
We use two types of data for our experiments: public datasets and an additional colloected dataset. Our data preprocessing pipeline is similar to that of LatentSync~\citep{li2024latentsync}. The core steps include detecting and cropping face regions from raw videos, followed by filtering out clips with low SyncNet~\citep{Chung16a_SyncNet} scores to ensure data quality. All data are finally processed into talking-head-only videos at a resolution of $512 \times 512$, matching the input size of our model. We use two public datasets, HDTF~\citep{zhang2021hdtf_dataset} and Realtalk~\citep{geng2023theRealTalkData}. After processing, our open-source training set contains approximately 50k video clips, each with a duration of about 8 seconds.
Our colloected dataset contains approximately 100k clips of human faces, processed using the same pipeline. This dataset is used exclusively as an additional training source to improve model robustness and is not used for evaluation. The AE is trained on both datasets and the Audio-to-Motion Transformer is trained on Open-Source Datasets only. All videos are resampled to 25FPS.

\subsection{Evaluation of Speaker}
\label{sec:4.3}
We first compare all methods in offline version in Sec. \ref{sec:4.3}, \ref{sec:4.4} and \ref{sec:4.5}, \textit{i.e.}, the audio feature are encoded in once via audio encoder (the encoder could see the future audio). Then discuss the online version in Sec \ref{sec:4.6}.

We focus on automated, no-reference metrics for evaluation in this subsection. Metrics that require a ground-truth video are discussed in our ablation studies. Following the evaluation protocol of recent video generation benchmarks~\citep{huang2024vbench}, we evaluate our method across four key dimensions: lip-sync quality~\citep{Chung16a_SyncNet}, content similarity~\citep{radford2021learning}, image quality~\citep{huang2024vbench}, motion dynamics~\citep{huang2024vbench}. The specific metrics used to quantify each dimension are detailed in Appendix.

We compare \ourmethod with several representative or state-of-the-art talking-head generation models:
SadTalker~\citep{zhang2023sadtalker}, Real3DPortrait~\citep{ye2024real3d}, 
Hallo3~\citep{cui2024hallo3}, Sonic~\citep{ji2025sonic}, and our reproduced INFP~\citep{zhu2025infp}. For evaluation, we construct a unified test set by randomly sampling $100$ clips each from the HDTF and our internal datasets, these clips were filtered out from training. All input faces are cropped and resized to $512 \times 512$.

To ensure a fair and comprehensive comparison, we establish two additional baselines specifically designed for frame-by-frame generation.
The first, which we denote as INFP-I (INFP-Inpainting), is an adaptation of our reproduced INFP baseline. We modify its chunk-by-chunk framework: for a fixed chunk size $k = 32$, we decrease the number of generated frames to $n$ and use $k-n$ frames as past context. We tested $n \in \{1, 2, 4, 8\}$, which corresponds to latencies of $\{40, 80, 160, 320\}$\,ms, respectively, in our 25\,FPS setting.
It is important to note that this strategy must be consistently applied during both training and inference; applying it only at inference time to a model trained on full chunks results in severe temporal instability and jitter.
The second baseline is a deterministic ablation of our own proposed method. We achieve this by removing the flow matching head from our architecture. This transforms our probabilistic model into a deterministic one that autoregressively predicts the motion frame by frame.

\begin{figure}
    \centering
    \includegraphics[width=\linewidth]{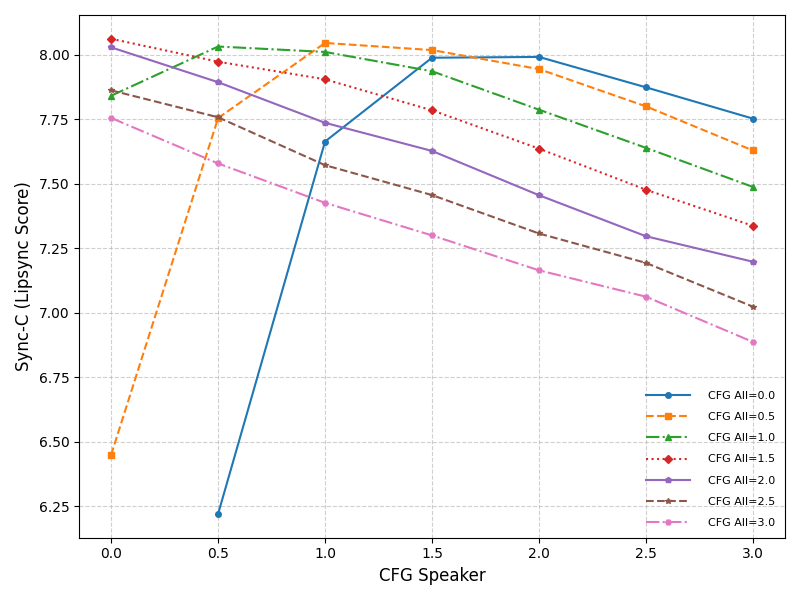}
    \caption{Effect of classifier-free guidance (CFG) on lipsync performance. We evaluate Sync-C across a grid of CFG All and CFG Speaker.}
    \label{fig:cfg}
\end{figure}

The lip-sync performance is mainly determined by the guidance weights $w_{all}$ and $w_{s}$. As shown in Figure~\ref{fig:cfg}, we empirically found that optimal performance is typically achieved when their sum, $w_{all} + w_{s}$, is approximately 1.5. Therefore, we set $w_{all}=1.0$ and $w_{s}=0.5$ as the chosen hyperparameters for our model.

\begin{table}[t]
    \centering
    \caption{\textbf{Comparison with existing methods.} 
    We compare our method with state-of-the-art video diffusion models (\textit{slower}) and other warp-based methods (\textit{faster}) on the HDTF-100 test set. $*$ denotes our reproduced version. Latency means the reaction time to the most recent incoming audio.}
    \label{tab:speaker}
    \resizebox{\linewidth}{!}{
    \begin{tabular}{l cccc}
        \toprule
        & {Sync-C}$\uparrow$ & {CS}$\uparrow$ & {Quality}$\uparrow$ & {Dynamic}$\uparrow$ \\
        \midrule
        \textcolor{gray}{GroundTruth} 
        & \textcolor{gray}{7.614} & \textcolor{gray}{0.928} & \textcolor{gray}{0.642} & \textcolor{gray}{0.660} \\
        Hallo3~\citep{cui2024hallo3}  
        & 6.814 & 0.915 & 0.638 & \textbf{0.870} \\
        Sonic~\citep{ji2025sonic} 
        & \textbf{8.495} & 0.935 & 0.626 & 0.600 \\
        \midrule
        SadTalker~\citep{zhang2023sadtalker} 
        & 6.704 & \textbf{0.965} & \textbf{0.697} & 0.080 \\
        Real3DPortrait~\citep{ye2024real3d} 
        & 6.811 & 0.943 & 0.637 & 0.030 \\
        \midrule
        INFP*~\citep{zhu2025infp} 
        & 6.894 & 0.909 & 0.643 & 0.950 \\
        INFP\_I* (latency $40$ ms) 
        & 1.718 & 0.877 & 0.630 & 0.580 \\
        INFP\_I* (latency $80$ ms)  
        & 3.112 & 0.897 & 0.655 & 0.270 \\
        INFP\_I* (latency $160$ ms)
        & 5.335 & 0.939 & 0.644 & 0.040 \\
        INFP\_I* (latency $320$ ms)
        & 6.637 & 0.945 & 0.638 & 0.210 \\
        \midrule
        {Ours}
        & \uline{8.136} & \uline{0.947} & \uline{0.642} & \uline{0.580} \\
        {Ours (w/o Flow Matching Head)}
        & {7.867} & 0.940 & 0.643 & 0.680 \\
        {Ours (w/o Frame Wise Addition)}
        & 7.660 & 0.925 & 0.648 & 0.470 \\
        \midrule
        Ours (latency $40$ ms)  
        & 6.896 & 0.937 & 0.647 & 0.570 \\
        Ours (latency $80$ ms)
        & 7.671 & 0.959 & 0.645 & 0.520 \\

        \bottomrule
    \end{tabular}}
\end{table}

\subsection{Evaluation of Listener}
\label{sec:4.4}
To quantitatively evaluate the listener performance, we conduct evaluations on the RealTalk test set, which comprises $38$ videos ranging from $3$ to $14$ seconds in length. We compare our method against our own reproduced of INFP. The evaluation relies on facial expression coefficients (Exp) and head poses (Pose) extracted using MediaPipe~\cite{lugaresi2019mediapipe}. We employ four key metrics: Fréchet Distance (FD) and Mean Squared Error (MSE) to measure the realism and accuracy of the generated motions, and two metrics to evaluate diversity: Variance (Var) and SI Diversity (SID). For SID, we follow the methodology of~\cite{ng2022learning2listen}, which assesses the diversity of facial and head motions by applying k-means clustering (with $k = 40$ for expression and k=20 for rotation) on ground truth and then calculating the entropy of the resulting histogram of cluster IDs within each sequence.

\begin{table}[t]
\centering
\renewcommand{\arraystretch}{1.1}
\setlength{\tabcolsep}{2pt}
\small
\caption{Quantitative comparison on FD, MSE, SID and Var for expression (Exp) and pose (Pose). Lower FD/MSE and higher SID/Var are better.}
\resizebox{\columnwidth}{!}{
\begin{tabular}{lcccccccc}
\toprule
\multirow{2}{*}{Method} &
\multicolumn{2}{c}{FD$\downarrow$} &
\multicolumn{2}{c}{MSE$\downarrow$} &
\multicolumn{2}{c}{SID$\uparrow$} &
\multicolumn{2}{c}{Var$\uparrow$} \\
\cmidrule(lr){2-3} \cmidrule(lr){4-5} \cmidrule(lr){6-7} \cmidrule(lr){8-9}
& Exp & Pose & Exp & Pose & Exp & Pose & Exp & Pose \\
\midrule
INFP~\cite{zhu2025infp} & 0.141 & \textbf{3.158} & 0.019 & \textbf{1.286} & 4.263 & 2.562 & 0.185 & 0.200 \\
\midrule
Ours  & 0.074 & 3.192 & \textbf{0.018} & 1.636 & 4.586 & \textbf{3.070} & \textbf{0.275} & \textbf{0.596} \\
Ours (w/o fm) & 0.079 & 3.236 & 0.020 & 1.472 & 4.570 & 2.896 & 0.246 & 0.556 \\
Ours (80ms) & \textbf{0.069} & 3.489 & 0.019 & 1.503 & \textbf{4.602} & 2.831 & 0.263 & 0.338 \\
\bottomrule
\end{tabular}
}
\label{tab:listner}
\vspace{-0.3em}
\end{table}

\subsection{Ablation Study}
\label{sec:4.5}
\paragraph{Feature Frame Wise Addition.}
We inject dyadic audio and anchor information via direct, frame-wise addition along the temporal dimension, instead of the cross-attention mechanism. As shown in Tables~\ref{tab:speaker} and ~\ref{tab:inferperformance}, this approach yields better lip-sync performance. 
The suggests for this task, the audio feature within a local window is enough for lip-syncing. 
Furthermore, by replacing computationally expensive cross-attention with simple addition, we achieve a 23.1\% increase in inference speed.

\begin{figure*}[t]
\begin{center}
\includegraphics[width=\textwidth]{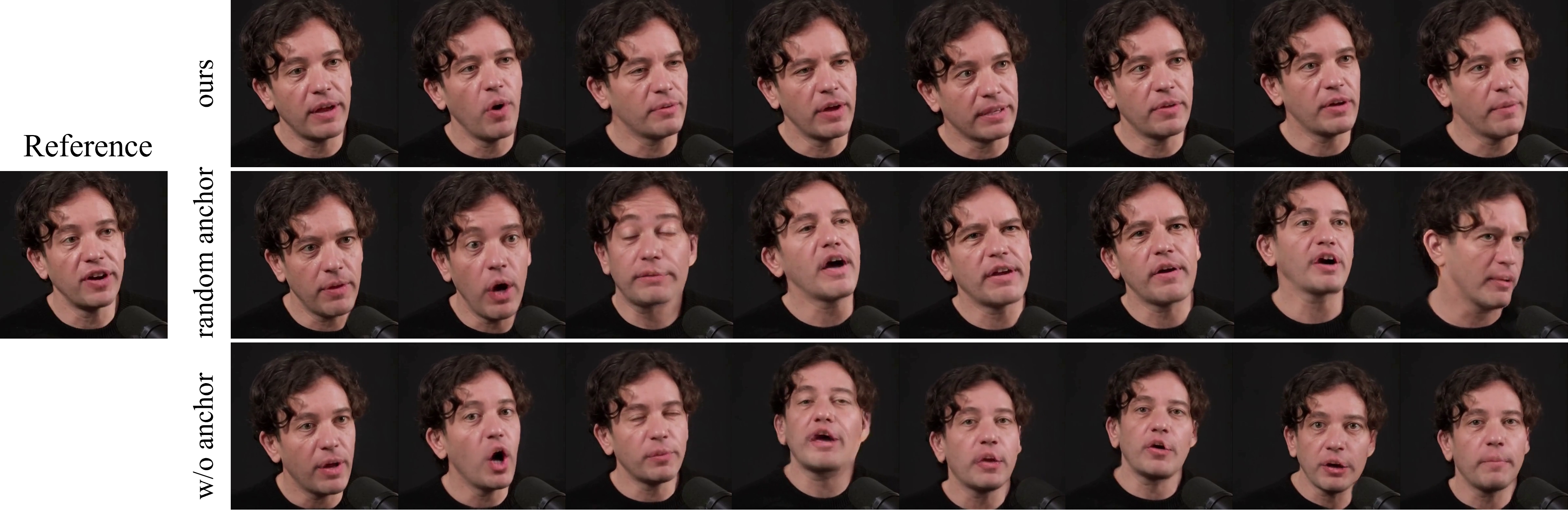}
\end{center}
\vspace{-0.3cm}
\caption{\textbf{Ablation Study on Anchor Frame Selection Strategy.} 
Visual comparison of long-sequence generation results using the same input audio and initial reference image. The rows from top to bottom correspond to three different anchor selection strategies used during training: our method (anchor sampled from the last 10 frames), a random anchor, and no anchor.}
\vspace{-0.3cm}
\label{fig:anchor}
\end{figure*}

\paragraph{Anchor Frame Selection.}
The anchor frame is crucial for mitigating pose drift in long-sequence autoregressive generation. We designed an ablation study with three settings: (1) our method, where the anchor is randomly sampled from the last 10 frames of the training sequence, (2) an anchor frame randomly sampled from the entire sequence, and (3) no anchor. As shown in Figure~\ref{fig:anchor}, our approach proves most effective at preventing pose drift over long generation periods.

\paragraph{Flow Matching Head.}
By removing the flow matching head, our model is converted into a deterministic autoregressive model. As detailed in Tables~\ref{tab:speaker} and~\ref{tab:listner}, the listener mode shows a notable decrease in motion diversity, as the model loses its ability to generate varied, stochastic reactions. In contrast, the speaker mode's performance degrades only marginally. We hypothesize this is because the listener mode is a more pronounced one-to-many mapping, where stochastic modeling provides a clear benefit.

\begin{table}[t]
\centering
\caption{\textbf{Comparison of different Wav encoders.} 
$\infty$ and $0$ denotes full-attention and pure-causal architecture, respectively. Top to bottom: (1) different pretrained full-attention wav encoder performs similar; (2) all pure-causal wav encoder show limited performance; (3) training with distillation from full-attention version could slightly improve the results; (4) lookahead improves the performance of Rope-Wav2vec2 (w. distill) (5) training Rope-Wav2vec2 with VAE 
style generation loss have a limited improvement, suggest that lookahead is simple yet effective.}
\vspace{0.1cm}
\small
\setlength{\tabcolsep}{4pt}
\begin{tabular}{lcc c}
\toprule
\multirow{2}{*}{Wav Encoder} & \multicolumn{2}{c}{Lookahead Time (ms)} & \multirow{2}{*}{{Sync-C} $\uparrow$} \\
\cmidrule(lr){2-3}
 & Train & Inference & \\
\midrule
Wav2vec2~\cite{baevski2020wav2vec} & $\infty$ & $\infty$ & 8.114 \\
Wav2vec2 (random init) &  $\infty$ &  $\infty$ & 4.568 \\
Wav2vec2 (freeze) &  $\infty$ &  $\infty$ & 6.037 \\
Rope-Wav2vec2 &  $\infty$ &  $\infty$ & 8.136 \\
Hubert~\cite{hsu2021hubert} &  $\infty$ &  $\infty$ & 8.007 \\
Wavlm~\cite{chen2022wavlm} &  $\infty$ &  $\infty$ & 7.719 \\

\midrule
Wav2vec~\cite{schneider2019wav2vec} & 0 & 0 & 1.761 \\
Mimi~\cite{defossez2024moshi} & 0 & 0 & 5.628 \\
Encodec(24KHz)~\cite{defossez2022encodec} & 0 & 0 & 2.034 \\
\midrule
Rope-Wav2vec2  & $\infty$ & $\infty$ & 8.136 \\
Rope-Wav2vec2 (w/o distill) &  0 &  0 & 2.861 \\
Rope-Wav2vec2 (w. distill) &  0 &  0 & 3.017 \\
\midrule
\multirow{5}{*}{Rope-Wav2vec (lookahead)} 
 & 0 & 0 & 3.017 \\
 & 20 & 20 & 5.976 \\
 & 40 & 40 & 6.896 \\
 & 60 & 60 & 7.387 \\
 & 80 & 80 & 7.671 \\
\midrule
Rope-Wav2vec2-VAE & 0 & 0 & 3.495 \\
\bottomrule
\end{tabular}
\label{tab:wavencoder}
\end{table}

\begin{table}
\centering
\small
\caption{\textbf{Inference performance.} Inference speeds measured on an H200 GPU, where {APD (Audio Packet Delay)} denotes the latency between the arrival of a new audio packet and the generation of the corresponding facial motion, and Step denotes the flow matching denoising step. Result of our model is with lookahead $60$ ms version.}
\begin{tabular}{lccccc}
\toprule
Method &Step & FPS & APD(s) & Sync-C$\uparrow$\\
\midrule
INFP & 5 & 21.27  &  1.0070 & 6.104     \\
INFP & 10 & 11.37  &    1.0479  & 6.381   \\
\midrule
Ours  & 1 & 38.46  & 0.0262 & 7.165\\
Ours & 5 & 29.24  & 0.0342 & 7.371\\
Ours & 10 & 21.88 & 0.0457 & 7.387\\
\midrule
Ours (w/o Add) & 5 & 22.47 & 0.0445 & 7.119 \\
\bottomrule
\end{tabular}
\label{tab:inferperformance}
\end{table}

\begin{figure*}[t]
\begin{center}
\includegraphics[width=1.0\textwidth]{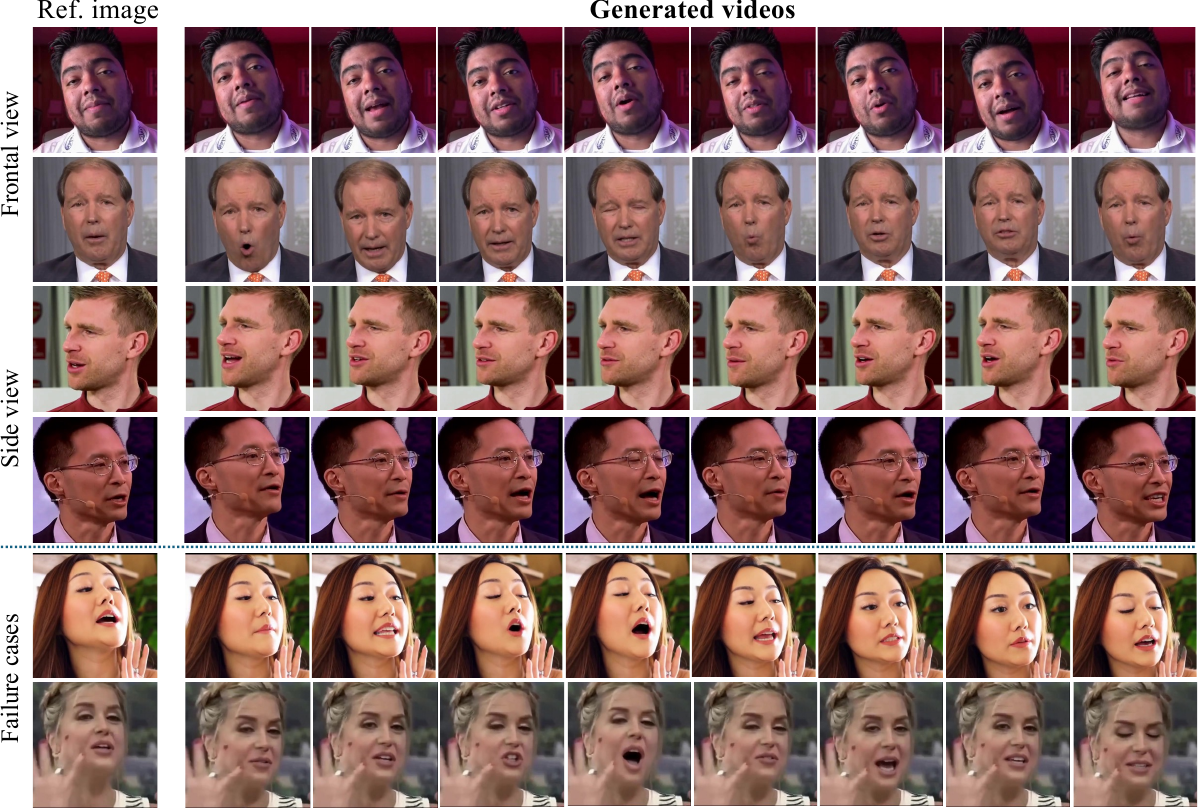}
\end{center}
\vspace{-0.3cm}
\caption{\textbf{Qualitative results and limitations.} \textit{Top:} Given reference images from different views, our approach can generate realistic videos that maintain consistent head pose over time. \textit{Bottom:} Our method have the limitation when hands are overlapped with face.   
}
\vspace{-0.3cm}
\label{fig:cases}
\end{figure*}

\subsection{Wav Encoder for Online Generation}
\label{sec:4.6}

We then discuss the online version, the selection of the audio encoder will influence the performance on downstream tasks. We conducted a comprehensive evaluation of several prominent audio encoders to determine their suitability for our task. The candidates included non-causal encoders: Wav2vec2 \citep{baevski2020wav2vec}, Hubert \citep{hsu2021hubert}, WavLM \citep{chen2022wavlm}, and causal encoders: Wav2vec \citep{schneider2019wav2vec}, Mimi \citep{defossez2024moshi}, Encodec~\cite{defossez2022encodec}, and our Rope-Wav2vec2.

As shown in the Table~\ref{tab:wavencoder}, group 1. our experiments provide several insights into the role of the audio encoder. First, we observe that wav2vec2, Hubert, and WavLM yield comparable performance, which we attribute to their similar network architectures and pre-training objectives. 
To emphasize the importance of pre-training, we also trained a model with a randomly initialized wav2vec2, which led to a substantial drop in performance. This confirms that a well-pre-trained audio encoder is critical for this task.
Furthermore, unlike INFP, which freezes the audio encoder during training, our method finetunes its parameters throughout the process. An ablation study where we froze the wav2vec2 encoder resulted in a significant performance degradation. This finding likely explains why our method outperforms INFP, even without resorting to more complex operations on the audio features, such as memory bank. The continuous fintuning of the encoder during training is the key factor in our model's better performance.

Then, we discuss the different design choices to make the audio encoder causal. 

\begin{enumerate}
    \item \textbf{Apply pretrained causal encoders directly.} Instread of using Wav2Vec2 \citep{baevski2020wav2vec}, we test wav encoder origianlly designed with causal architectures, such as Mimi~\cite{defossez2024moshi}, Encodec(24KHz)~\cite{defossez2022encodec} and Wav2vec \citep{schneider2019wav2vec}.
    \item \textbf{Train a causal encoder with distillation.} We modified our Wav2Vec2 encoder to make it causal, noted as Rope-Wav2vec2, and first train a the full-attention version, then distill a causal version.   
    \item \textbf{Train a causal encoder with lookahead.} We operate on the attention mask to allow Rope-Wav2vec2 could use a few future audio features.   
    \item \textbf{Train a generative causal encoder.} We apply generative model style training for the Rope-Wav2vec2, \textit{e.g.}, during the distillation phase, apply a VAE style loss. 
\end{enumerate}

As shown in Table~\ref{tab:wavencoder}, group $2-4$, 
The empirical evidence strongly suggests that for high-fidelity audio-driven talking head generation, a small window of future audio is not just beneficial but essential for correctly modeling the co-articulation and anticipatory dynamics of human speech.

\section{Limitations and Future Works}
Our method inherits some structural limitations from its warp-based foundation. First, inaccuracies in the warping process can occasionally lead to rigid deformations on accessories like jewelry or glasses. 
Second, it is challenging for our method to deal with hand occlusion. As shown in Figure~\ref{fig:cases}, the generated videos exhibit static hand motion. 
Additionally, our inference pipeline requires the input face to be cropped and aligned to a fixed position and scale, similar to the training data. Finally, while our autoregressive model can access historical context, it lacks a dedicated design to ensure long-term diversity. This is largely due to the scarcity of continuous, long-form training data, presenting a key challenge we will explore in the future.

{
    \small
    \bibliographystyle{ieeenat_fullname}
    \bibliography{main}
}
\newpage
\input{supp}
\end{document}

%% file: preamble.tex
\usepackage{ulem}



%% file: supp.tex
\clearpage
\setcounter{page}{1}
\maketitlesupplementary
\appendix

\noindent This supplemental document contains four sections: 
\begin{itemize}[leftmargin=*]

\item Video and Codes (Section \ref{sec:sup1}). 

\item Disentangled AE for Motion Representation (Section \ref{sec:sup2}).

\item Inference Performance (Section \ref{sec:sup3}).

\item Metric Definitions (Section \ref{sec:sup4}).

\end{itemize}

\section{Video and Codes}
\label{sec:sup1}

This supplemental provides a comprehensive HTML that contains separate videos, including:

\begin{itemize}
    \item Results for dyadic talking heads generation.
    \item Comparison between our method and INFP baseline. 
    \item Ablation study for the lookahead module and anchor. 
    \item More results for in-the-wild image and audios.
\end{itemize}

\section{Disentangled AE for Motion Representation}
\label{sec:sup2}
\paragraph{Architecture.} 
The VAE is an image-level autoencoder that disentangles a source image $I_s$ into an appearance feature $\mathbf{v}_{app}$ and a motion feature $\mathbf{m}$ in the latent space.
To enhance the robustness of facial representations, we leverage a 3D-aided appearance encoder ($E_{app}$) and a face decoder ($D_{vae}$) with the same network architecture as LivePortrait~\citep{guo2024liveportrait}. 
The 3D appearance feature volume provides a more accurate characterization of facial appearance in 3D compared to traditional 2D feature maps. 
We also employ a 2D-based motion encoder ($E_{m}$) to extract motion codes and compress them into low-dimensional 1D representations. 
This design effectively captures the core semantics of facial motion while avoiding entanglement with appearance information~\citep{wang2023progressive, wang2024lia}.

Following LIA, we build a motion flow-based generation pipeline.
Concretely, $E_{m}$ first extracts the source motion code $\mathbf{m_{s}}$ from the input portrait image $I_s$, and driving motion codes $\mathbf{m^{1:N}_{dri}}$ from the driving video $V_{dri}$. 
A motion flow estimation model $F$ then takes these motion codes as input to predict the \textit{self$\rightarrow$driving} flow $\mathit{Flow}_{s \rightarrow d}$.
We utilize the $E_{app}$ to extract a 3D appearance feature volume ($\mathbf{v}_{app}$) from the portrait image $I_s$. 
$\mathbf{v}_{app}$ is then warped using the estimated motion flow and passed to the decoder $D_{vae}$ to synthesize the final predicted video $I^{1:N}_{pred}$.
The overall pipeline is formulated as follows:
$$
\mathbf{m_{s}}=E_{m}(I_s); \mathbf{m^{1:N}_{dri}}=E_{m}(V_{dri})
$$
$$
\mathbf{v}_{app}=E_{app}(I_s)
$$
$$
\mathit{Flow_{s \rightarrow d}}=F(\mathbf{m_{s}}, \mathbf{m^{1:N}_{dri}})
$$
$$
I^{1:N}_{pred}=D_{vae}(Warp(\mathbf{v}_{app}, \mathit{Flow_{s \rightarrow d}}))
$$
Readers are referred to~\citep{wang2024lia} for more details of this architecture.

\paragraph{Training Objective.} 
The VAE is trained with a composite loss function. 
We adopt the base losses from LIA~\citep{wang2024lia}, which include reconstruction ($\mathcal{L}_{rec}$), perceptual ($\mathcal{L}_{feat}$), and adversarial ($\mathcal{L}_{adv}$) losses. 
We adopt two additional losses following~\citep{drobyshev2022megaportraits}: a cycle consistency loss ($\mathcal{L}_{cyc}$) to prevent appearance leakage and enhance the disentanglement between motion and appearance features; and a gaze direction loss ($\mathcal{L}_{gaze}$) to encourage diversity in eye movement.
To further enhance the disentanglement between motion and appearance features, we follow MegaPortrait~\citep{drobyshev2022megaportraits} and incorporate severl additional losses: a cross-identity consistency loss ($\mathcal{L}_{cyc}$), an identity consistency loss ($\mathcal{L}_{id}$)
a gaze direction loss ($\mathcal{L}_{gaze}$). 
We additionally incorporate a gaze direction loss ($\mathcal{L}_{gaze}$) to encourage diversity in eye movement.
The final objective is a weighted sum of these terms:
\begin{equation}
\resizebox{0.9\linewidth}{!}{%
    $\mathcal{L}_{VAE} = \lambda_{rec}\mathcal{L}_{rec} + \lambda_{adv}\mathcal{L}_{adv} + \lambda_{feat}\mathcal{L}_{feat} + \lambda_{cyc}\mathcal{L}_{cyc} + \lambda_{gaze}\mathcal{L}_{gaze}$%
}
\end{equation}
The weights $\lambda$ for each term are determined empirically ($\lambda_{rec}=1$, $\lambda_{feat}=2$, $\lambda_{adv}=0.2$, $\lambda_{gaze}=5$, $\lambda_{cyc}=1$).

\section{Inference Performance}
\label{sec:sup3}
In Table \ref{tab:inferperformance}, we report that our method achieves a FPS of 29.24 and an APD (Audio Packet Delay) of 0.0342 with a flow matching step of 5. 
By reducing the data offload between the CPU and GPU, we could archive an FPS of 38.61 and an APD of 0.0259 on the same server.

Furthermore, we conduct tests on a consumer-grade host equipped with a consumer-grade GPU (4090) and an i9-13900KF CPU. On this setup, our method achieves a FPS of 58.48 and an APD of 0.0171. The entire process consumed only 2.37G of VRAM.

\section{Head Pose Stability (Anchor)}
To quantitatively evaluate the anti-drifting efficacy, we measure the 6-DoF head pose drift relative to the initial frame using MediaPipe. We report \textbf{Mean/Max Drift} (average/maximum deviation) and \textbf{Std Drift} (fluctuation stability).
As shown in Table~\ref{tab:head_pose_drift}, we tested on 10 audio clips (3 minutes each) driven by 10 different reference images.
Our method demonstrates exceptional stability across all durations (10s, 60s, 180s). Specifically, our rotation drift remains under $8^\circ$ even after 3 minutes, significantly outperforming the ``Random Anchor'' and ``No Anchor'' baselines, which suffer from severe cumulative error (drifting $>34^\circ$).

\begin{table}[t]
\centering
\caption{Head Pose Drift Evaluation. We evaluate head pose stability across three different anchor strategies. The metrics reported for Rotation are in degrees, and for Position are in relative units from MediaPipe. Lower values indicate better stability.}

\label{tab:head_pose_drift}

\renewcommand{\arraystretch}{1.2}

\resizebox{\columnwidth}{!}{
\begin{tabular}{c l ccc c ccc}
\toprule
\multirow{2}{*}{Duration} & \multirow{2}{*}{Method} &
\multicolumn{3}{c}{Rotation (degrees)} & & \multicolumn{3}{c}{Position} \\
\cmidrule(lr){3-5} \cmidrule(lr){7-9}

 & & Mean Drift $\downarrow$ & Max Drift $\downarrow$ & Std Drift $\downarrow$ & &
   Mean Drift $\downarrow$ & Max Drift $\downarrow$ & Std Drift $\downarrow$\\
\midrule

\multirow{3}{*}{10s} 
 & Ours     & $\mathbf{6.61}$  & $\mathbf{12.48}$ & $\mathbf{2.25}$ & & $\mathbf{1.04}$ & $\mathbf{2.45}$ & $\mathbf{0.53}$ \\
 & Random Anchor & $11.65$ & $20.76$ & $3.83$ & & $1.62$ & $3.32$ & $0.74$ \\
 & No Anchor    & $30.12$ & $41.08$ & $8.47$ & & $2.19$ & $4.17$ & $0.84$ \\

\midrule

\multirow{3}{*}{60s}
 & Ours     & $\mathbf{7.73}$  & $\mathbf{17.61}$ & $\mathbf{2.45}$ & & $\mathbf{1.16}$ & $\mathbf{3.26}$ & $\mathbf{0.57}$ \\
 & Random Anchor & $12.69$ & $24.27$ & $3.49$ & & $1.62$ & $3.87$ & $0.76$ \\
 & No Anchor    & $33.96$ & $43.25$ & $4.59$ & & $2.38$ & $4.56$ & $0.73$ \\

\midrule

\multirow{3}{*}{180s}
 & Ours     & $\mathbf{7.62}$  & $\mathbf{19.31}$ & $\mathbf{2.72}$ & & $\mathbf{1.20}$ & $\mathbf{3.69}$ & $\mathbf{0.59}$ \\
 & Random Anchor & $12.86$ & $26.90$ & $4.62$ & & $1.62$ & $4.09$ & $0.75$ \\
 & No Anchor    & $34.52$ & $44.45$ & $3.31$ & & $2.42$ & $4.68$ & $0.71$ \\

\bottomrule
\end{tabular}
}
\end{table}

\section{Metric Definitions}
\label{sec:sup4}
\paragraph{Lip-sync Consistency (Sync-C):} 
We employ the confidence score derived from SyncNet~\cite{Chung16a_SyncNet}, referred to as SyncScore, to evaluate the temporal synchronization quality between the generated lip movements and the corresponding audio signals. Specifically, we extract audio features and visual features from consecutive video frames, and compute the cross-correlation between audio and visual embeddings across multiple temporal shifts. The confidence score represents the maximum correlation value, indicating the degree of audio-visual synchronization. The final metric is computed as a weighted average across all frames, where higher values indicate better lip-sync quality.


\paragraph{Content Similarity (CS):} 
To evaluate the temporal consistency of the generated content, we employ Content Similarity metric. Specifically, we extract deep visual features from all video frames using a pre-trained encoder, resizing each frame to 160×160 pixels. The first frame serves as the reference, and we compute the cosine similarity between its feature embedding and those of all subsequent frames. The final score is the mean cosine similarity across all frame pairs, with values closer to 1 indicating higher temporal consistency.

\paragraph{Imaging Quality (Quality):} 
We assess the visual quality of generated videos using a pre-trained image quality assessment model. Each frame in the video is independently evaluated by the model, producing a quality score. The Imaging Quality metric is computed as the average score across all frames, normalized to the range [0, 1], where higher values indicate superior visual fidelity and perceptual quality.

\paragraph{Dynamic Degree (Dynamic):} 
To quantify the presence of motion in generated videos, we utilize the Dynamic Degree metric based on optical flow analysis. We employ the RAFT model~\cite{raft} to estimate inter-frame optical flow, sampling frames at intervals of fps/8. For each consecutive frame pair, we compute the flow magnitude and take the mean of the top 5\% largest magnitudes as the motion score. A video is classified as dynamic if the number of frames exceeding an adaptive threshold (scaled by resolution and video length) surpasses a minimum count threshold. The final metric represents the proportion of videos exhibiting significant motion.

\paragraph{Fr\'echet Distance (FD):} 
Motion realism is quantified by measuring the distributional discrepancy between the generated motion sequences and the ground-truth motions. Specifically, we calculate the Fr\'echet Distance (FD) across the feature domains of facial expressions and head poses over the entire sequence.

\paragraph{Mean Squared Error (MSE):} 
We calculate the Mean Squared Error (MSE) between the generated motion parameters and the ground-truth motion sequences to evaluate the deterministic reconstruction accuracy.

\paragraph{Shannon Index for Diversity (SID):} 
Following L2L~\cite{ng2022learning2listen}, this metric quantifies the magnitude and dynamism of generated facial motions by calculating the variance along the temporal dimension. We extract facial motion representations using MediaPipe Face Landmarker, obtaining 6-dimensional head motion parameters (3D translation and 3D rotation in Rodrigues form) and 52-dimensional expression blendshapes for each frame. The variance is computed separately for head motion ($\text{Var}_{\text{head}}$) and facial expressions ($\text{Var}_{\text{exp}}$) as $\text{Var} = \sum_{d=1}^{D} \text{Var}(\mathbf{x}^{(d)})$, where $\mathbf{x}^{(d)}$ denotes the motion sequence along dimension $d$. Higher variance indicates more dynamic and expressive motions.

\paragraph{Variance (Var):} 
Following L2L~\cite{ng2022learning2listen}, this metric quantifies the magnitude and dynamism of generated facial motions by calculating the variance along the temporal dimension. We extract facial motion representations using MediaPipe Face Landmarker, obtaining 6-dimensional head motion parameters (3D translation and 3D rotation in Rodrigues form) and 52-dimensional expression blendshapes for each frame. The variance is computed separately for head motion ($\text{Var}_{\text{head}}$) and facial expressions ($\text{Var}_{\text{exp}}$) as $\text{Var} = \sum_{d=1}^{D} \text{Var}(\mathbf{x}^{(d)})$, where $\mathbf{x}^{(d)}$ denotes the motion sequence along dimension $d$. Higher variance indicates more dynamic and expressive motions.

\paragraph{Audio Packet Delay (APD):}
This metric evaluates system latency in real-time streaming generation scenarios, defined as the time interval between the reception of an audio packet and the output of the corresponding motion. It is important to note that chunk-based methods, such as INFP, inherently require the accumulation of a full audio chunk before motion synthesis can commence. Consequently, the calculation of APD must encompass not only the model's computational inference time but also the buffering latency equivalent to the duration of the audio chunk. In our implementation of INFP, this chunk duration is set to 0.96 seconds.
